\documentclass{article}
\usepackage{spconf,amsmath,epsfig}
\usepackage{hyperref}
\usepackage{graphicx}
\usepackage{amsfonts}
\usepackage{amsmath}
\usepackage{upgreek}
\usepackage{multirow}
\usepackage{booktabs}
\usepackage[table,xcdraw]{xcolor}
\usepackage{tabularx}
\usepackage{caption} 
\title{SpACNN-LDVAE: Spatial Attention Convolutional Latent Dirichlet Variational Autoencoder for Hyperspectral Pixel Unmixing}

\twoauthors
 {Soham Chitnis\sthanks{Work done in MITACS Globalink Research Internship}}
	{Birla Inst. of Tech. \& Sci. Pilani\\
	Department of CS \& IS \\ 
	Goa, India}
 {Kiran Mantripragada, Faisal Z. Qureshi}
	{Ontario Tech University\\
	Faculty of Science\\
	Oshawa, Canada}
\begin{document}
%
\maketitle
\begin{abstract}
The hyperspectral pixel unmixing aims to find the underlying materials (endmembers) and their proportions (abundances) in pixels of a hyperspectral image.  This work extends the Latent Dirichlet Variational Autoencoder (LDVAE) pixel unmixing scheme by taking into account local spatial context while performing pixel unmixing.  The proposed method uses an isotropic convolutional neural network with spatial attention to encode pixels as a dirichlet distribution over endmembers.  We have evaluated our model on Samson, Hydice Urban, Cuprite, and OnTech-HSI-Syn-21 datasets. Our model also leverages the transfer learning paradigm for Cuprite Dataset, where we train the model on synthetic data and evaluate it on the real-world data.  The results suggest that incorporating spatial context improves both endmember extraction and abundance estimation. 
\end{abstract}
\section{Introduction}
\label{sec:intro}

Hyperspectral images can be applied in a variety of applications and are very prominent in the field of remote sensing. They are known for having lower spatial resolution but very high spectral resolution. The lower spatial resolution implies that a single pixel covers a large region of space~\cite{heylen_rob}. Such a phenomenon is mostly attributed to the images collected from high altitudes. The pixel unmixing task is to infer the materials with their mixing ratios. This problem is of particular interest to the hyperspectral image analysis community. In literature, it is sometimes referred to as \textit{spectral unmixing}~\cite{maggiori}. It involves recovering the ``pure'' spectra of materials known as endmembers with the ratios of mixing known as abundances.

Existing approaches can usually be divided into two classes: physics-based or data-driven methods.  Physics-based methods use phenomenological models for the radiance response of the materials~\cite{heylen_rob,Plaza2012}.  These are costly to implement, since it is often tedious to develop accurate radiance response models.  On the other hand, data-driven methods are easier to use in practice; however, these require access to data for model setup and training.  It is not always straightforward to collect this data.  Some examples of data-driven techniques are Blind Source Separation (BSS), Non-Negative Matrix Factorization (NMF), Principle Component Analysis (PCA), and Linear Discriminant Analysis (LDA)~\cite{nmf1, Khajehrayeni2021, Zhao2020, Plaza2005}.  A number of commonly used approaches, such as N-FINDR, Pixel Purity Index (PPI), and Vector Component Analysis (VCA), first extract endmembers and then estimate abundances~\cite{n-findr,ppi,vca} by leveraging abundances-sum-to-one (ASC) and abundances-non-negative (ANC) constraints~\cite{fcls}. 

Variants of (Linear) Mixing Models (LMM) include Perturbed LMM, Extended LMM, Non-linear LMM, and Data-driven LMM~\cite{ELMM1, ELMM2, PLMM, NLMM, DataLMM}. Many NMF methods---SSWNMF, SGSNMF, TV-RSNMF, GLNMF, $L_{1/2}$NMF---use regularization  achieve better performance~\cite{SSWNMF,SGSNMF,TV-RSNMF, GLNMF, L1/2NMF}. Recently, a number of deep learning approaches have been proposed to address the problem of hyperspectral pixel unmixing---EACNN, DAEN, DeepGUn, and TANet~\cite{EACNN, DAEN, DeepGUn, TANet}.

Data-driven methods can further be divided into supervised and unsupervised schemes. Unsupervised schemes, e.g., BSS, do not require access to labeled data for model setup and training. Supervised schemes, such as EACNN, on the other hand, require access to labeled data for training. The method proposed here requires access to labeled data. It is, however, possible to apply this method in scenarios where labeled data is not available by training the model on synthetic data and leveraging {\it transfer learning}.  

This paper builds on the work of the Latent Dirichlet Variation Autoencoder (LDVAE), which assumes that the endmember spectrum can be represented using a multivariate Normal Distribution and the mixing ratios (abundances) can be represented using Dirichlet distribution~\cite{ldvae}. LDVAE is a Multilayer Perceptron model (MLP), and it does not incorporate spatial information.  The approach presented in this paper aims to exploit spatial coherence, i.e., nearby pixels often contain the ``same'' endembers and have ``similar'' abundances.  The encoder stage uses Isotropic CNN layers plus a spatial attention layer to capture the local spatial structure around a pixel as its latent representation is constructed.  Subsequently, similar to LDVAE, the decoder uses this latent representation for endmember extraction and abundance estimation.
In the interest of clarity, hereafter, we will refer to the original LDVAE approach as MLP-LDVAE~\cite{ldvae}.

\section{Methodology}
\label{sec:methodology}

Latent Dirichlet Allocation (LDA) \cite{LDA}, a popular scheme in natural language processing, inspired MLP-LDVAE.  LDA assumes a collection of documents, where each document is a mixture of a set of topics.   Both the topics and their mixing ratios within each document is not known {\it a priori}.  This is similar to the problem of hyperspectral pixel unmixing.  A hyperspectral image is composed of pixels, and each pixel is a mixture of endmembers.  Both the endmembers and the abundances, i.e., the mixing ratios of endmembers, in any pixel are not known.  In the parlance of LDA, endmember extraction is akin to topic discovery and abundance estimation is similar to computing the mixing ratio of topics in a particular document.


    \begin{figure}
    \includegraphics[width=\columnwidth]{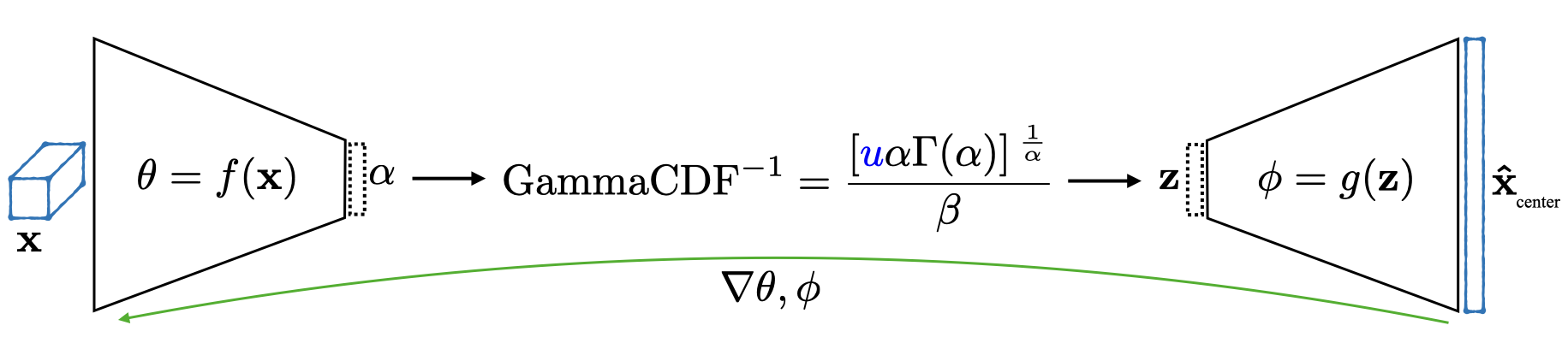}
    \caption{CNN Latent Dirichlet Variational Autoencoder.  Encoder $f$ takes an HSI patch $\mathbf{x}$ and constructs its latent representation (abundances).  The decoder stage is able to reconstruct the pixel spectrum given abundances.  Note that at training time the reconstruction loss is computed between the center pixel $\mathbf{x}_\text{center}$ and its reconstruction $\mathbf{\hat{x}}_\text{center}$.  } \label{fig1}
    \end{figure}
    
    Our model uses the VAE architecture depicted by Figure~\ref{fig1}. LDVAE is a Variational Autoencoder where the latent representation follows a Dirichlet distribution. The encoder is parameterized by $\theta$, which outputs the Dirichlet distribution parameter $\alpha$. LDVAE takes a single signal of the pixel. We now try to leverage spatial information in the encoder using a CNN.

\subsection{Spatial Attention Convolutional Neural Network Encoder}

    \begin{figure}
    \includegraphics[width=\columnwidth]{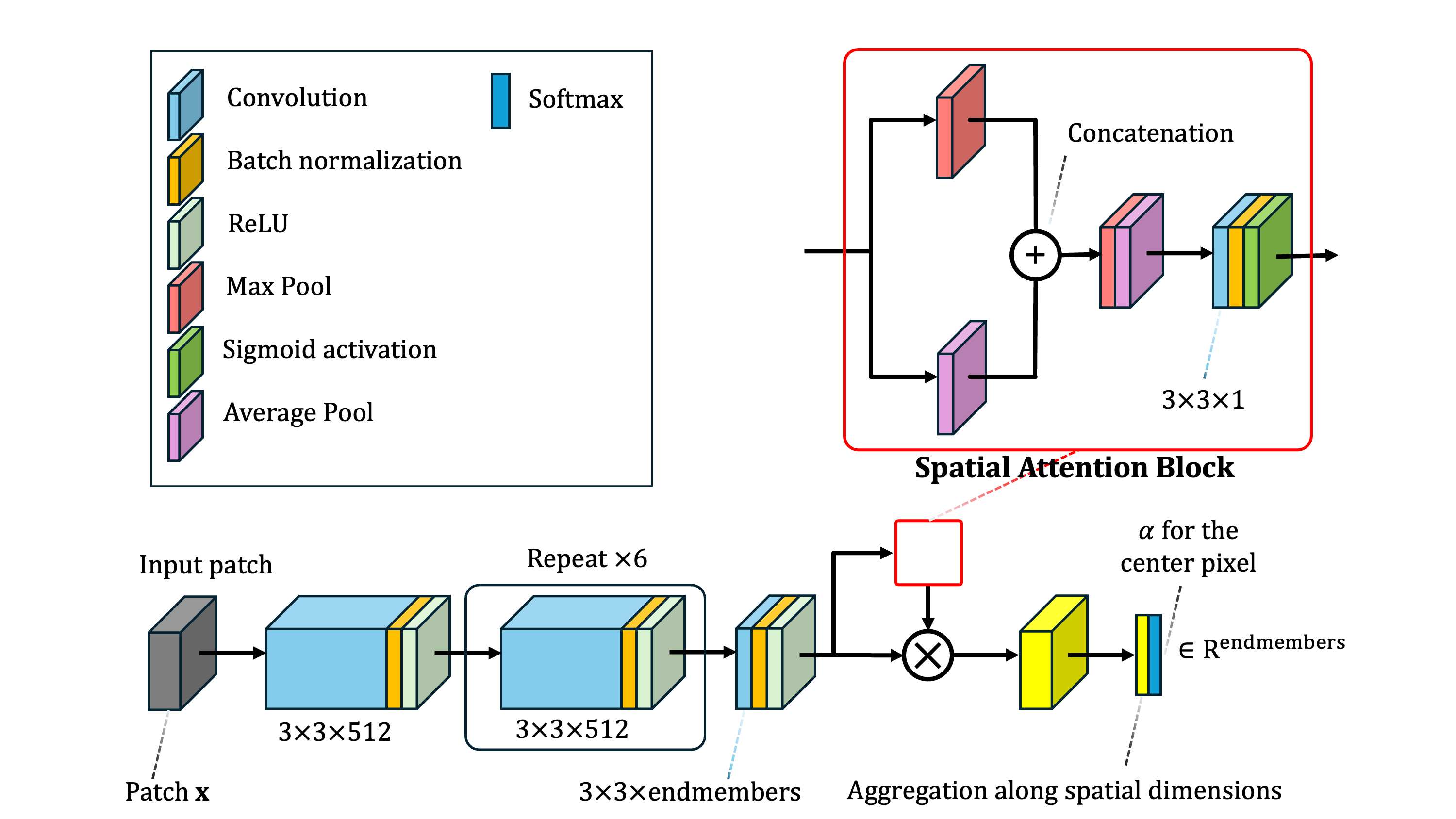}
    \caption{Spatial Attention Convolutional Neural Network Encoder.  The network takes an HSI patch $\mathbf{x}$ and returns abundances vector $\alpha$ for the center pixel $\mathbf{x}_\text{center}$.} \label{fig2}
    \end{figure}
    
    Our model uses a CNN encoder (Figure ~\ref{fig2}), which receives a rectangular patch as input and returns the Dirichlet distribution parameter $\alpha$ corresponding to the center pixel. The abundances $\mathbf{z}$ are sampled from the Dirichlet distribution and fed into the decoder, which reconstructs the spectral signal of the center pixel $\mathbf{\hat{x}}_\text{center}$.  The decoder follows the model used in~\cite{ldvae}.

    The encoder employs an isotropic CNN model; therefore, the spatial resolution is maintained.  The CNN encoder comprises of three modules: (1) stem, (2) body and (3) the spatial attention branch. The stem consists of a 2D convolution layer (kernel size: 3 and padding: 1) with Batch Normalization (BN) and Rectified Linear Unit (ReLU) activation.  The body consists of six blocks of convolution layer followed by BN and ReLU.  The spatial attention block follows the model introduced in~\cite{woo2018cbam}.  The output features are aggregated to generate the Dirichlet parameter $\alpha$. 

    Given the intermediate feature map $\mathbf{F} \in \mathbb{R}^{H\times W\times C}$, apply average and max pooling along the channel dimension.  Concatenate the results and perform 2D convolution with kernel size of 3 and apply sigmoid activation to obtain the 2D spatial attention map 
    \begin{equation}
        \mathbf{A} = \sigma{ \left( f^{3\times3} \left( \operatorname{AvgPool}_C(\mathbf{F}) \oplus \operatorname{MaxPool}_C(\mathbf{F}) \right) \right) },
    \end{equation}
    where $\sigma$ denotes the sigmoid function, $f^{3\times3}$ denotes the convolution operation with a $3 \times 3$ filter, $\oplus$ denotes the concatenation operation, and $\mathbf{A} \in \mathbb{R}^{H \times W}$.   The latent representation $\mathbf{z}' \in \mathbb{R}^C$ is computed as follows:
    \begin{equation}
       \mathbf{z}' = \sum_{i=1}^H \sum_{j=1}^W \mathbf{A}_{i,j} \mathbf{F}_{i,j}.
    \end{equation}
    Softmax activation is applied to obtain $\alpha$ in the final layer of encoder as follows:
    \begin{equation}
        \alpha_k = \frac{e^{z_k}}{\sum_{k=1}^C e^{z_k}}.
    \end{equation}
    This ensures that the model satisfies ASC and ANC. 
    
    \subsection{Spectral Reconstruction With Multivariate Normal Distribution}
    
    As stated previously, we employ the decoder used in MLP-LDVAE, which uses an MLP to 
     reconstructs the spectrum given abundances $\mathbf{z}$. The decoder serves two roles: (1) it is able to construct spectrum for previously unseen abundances and (2) it is able to perform endmember extraction by setting up the abundances appropriately.  The model assumes that spectra follow a multivariate Normal Distribution as below:
    \begin{equation}
        \begin{aligned}
            \mathbf{x} \sim Normal(\mathbf{x};\mu, \Upsigma)  \text{ where}\\
            \mathbf{x} = \{x_1, x_2, x_3, ..., x_k\} \\
            \mu = \{\mu_1, \mu_2, \mu_3, ..., \mu_k\} \\
            \Upsigma = \text{diag}(\sigma_1, \sigma_2, \sigma_3, ... \sigma_k)
        \end{aligned}
    \end{equation}

    \subsection{Loss function}

    The loss function has two components: (1) the reconstruction loss 
    \begin{equation}
        \mathcal{L}(\mathbf{z},\hat{\mathbf{z}}) = (\mathbf{z} \boldsymbol{-} \hat{\mathbf{z}})^{2},
    \end{equation} 
    where $\mathbf{z}$ and $\hat{\mathbf{z}}$ are ground truth and estimated abundances, respectively,
    and (2) the ELBO Loss 
    \begin{equation}
        \mathcal{L}(\mathbf{x};\theta,\phi) = \mathbb{E}_{q_\theta}[\log p_{\phi} \boldsymbol{-} \operatorname{KL} \left( q_{\theta}(\mathbf{z} | \mathbf{x}) || p( \mathbf{z} ) \right) ].
    \end{equation}
    The first term on the left side is the reconstruction loss for the spectrum.  The second term represents KL divergence that forces the latent representation towards a Dirichlet distribution.  For further details, please refer~\cite{ldvae,VAE,Blei_2017, Fox2012,kingma2022autoencoding}.

\section{Empirical Evaluation}

    \subsection{Datasets}
    All the experiments are conducted with four datasets. Samson is $95 \times 95$, $156$-channel hyperspectral image~\cite{Samson}. This dataset contains three endmembers: soil, tree, and water. UHYDICE Urban  is $307 \times 307$, $162$-channel hyperspectral image covering a $2 \times 2 \text{ m}^2$~\cite{Urban}. This dataset has three versions, containing four, five, and six endmembers. In this work, six endmembers are used. $80:20$ training/testing split is used for both datasets.  Cuprite dataset is a $512 \times 614$, $188$-channel hyperspectral image~\cite{Cuprite}. It contains twelve minerals (endmembers). This dataset does not provide ground truth abundances; therefore,  we use Cuprite-synthetic dataset for training~\cite{ldvae}.  This showcases transfer learning, where a model trained on synthetic dataset is subsequently used to perform inference on a real world dataset.  
    Lastly, OnTech-HSI-Syn-21 is a synthetic dataset containing nine endmembers. It is $128 \times 128$, $224$-channel hyperspectral image.
    
    \subsection{Metrics}
    
    {\bf Root Mean Square Error (RMSE)} is used to evaluate abundance estimation accuracy. It is computed as follows:
    $$
        \operatorname{RMSE} = \sqrt{\frac{1}{N}\sum_{n=1}^N (\mathbf{z_n} \boldsymbol{-} \mathbf{\hat{z}_n}) ^ 2},
    $$ where $\mathbf{z}$ is the ground truth abundances, and $\mathbf{\hat{z}}$ is generated abundances. $N$ denotes the number of pixels used in the computation.

    \noindent
    {\bf Spectral Angle Distance (SAD)} is a distance measurement between two spectral signals
        $$
            \operatorname{SAD} = \arccos {\left(\frac{\mathbf{\hat{x}_e ^ T}\mathbf{x_e}}{\|\mathbf{\hat{x}_e ^ T}\|\|\mathbf{x_e}\|}\right)},
        $$
    where $\mathbf{\hat{x}_e}$ represents the estimated endmember and $\mathbf{x_e}$ denotes the ground truth endmember. SAD is sometimes referred to as SAM~\cite{2014SpectralAM}.  In this work, we use SAD to capture endmember extraction accuracy. 
    
    \subsection{Experimental Settings}

     SpACNN-LDVAE was implemented using Pyro Python library~\cite{bingham2018pyro}.  All models were trained on NVidia V100-SXM2 gpus.  Adam optimizer was used with learning rate $0.001$~\cite{kingma2017adam}.

\section{Results}

    Table~\ref{table1} provides RMSE and SAD metrics (both endmember specific and average values) for Samson dataset.  Similarly Table~\ref{table2} lists these metrics for HYDICE Urban dataset.  Cuprite lacks abundance ground truth, so only endmember extraction results are provided for this dataset (see Table~\ref{table3}).  Tables~\ref{table4} and~\ref{table5} provide endmember extraction (SAD) and abundance estimation (RMSE) results for OnTech-Syn-HSI-21 dataset under various SNR settings, respectively.  The proposed method achieves lower RMSE and SAD numbers over MLP-LDVAE method.  However, the SAD scores on Cuprite are comparable to those achieved by MLP-LDVAE.  Recall that the model was trained on Cuprite Synthetic dataset, which lacks spatial coherence.  This is perhaps why the proposed method did not achieve better numbers than MLP-LDVAE.  This merits further discussion.  In all datasets, the proposed method achieves lower standard deviations.
    


\section{Conclusion}    

    This work extends MLP-LDVAE by replacing its encoder stage with a CNN with spatial attention.  This allows the encoder to attend to the neighbouring pixels when constructing the latent representation for a given pixel.  Our intuition is that hyperspectral images, similar to RGB images, exhibit spatial coherence, i.e. neighbouring pixels may contain similar endmembers and may exhibit similar abundances.   The proposed model is evaluated on four datasets and the results suggest that incorporating local spatial information improves both endmember extraction and abundance estimation.  Similar to MLP-LDVAE, our model is also capable of generating hyperspectral pixels given their abundances.

\begin{table*}[!h]
\centering
\caption{Abundance Estimation and Endmember Extraction Results on Samson Dataset}
\label{table1}
\resizebox{\textwidth}{!}{%
\begin{tabular}{@{}cccccccccc@{}}
\toprule
\multicolumn{2}{c}{} & SpACNN-LDVAE & MLP-LDVAE \cite{ldvae} & VCA+FCLS \cite{vca} & PLMM \cite{PLMM} & ELMM \cite{ELMM1, ELMM2} & GLMM \cite{GLMM} & DeepGUn \cite{DeepGUn} & EACNN \cite{EACNN} \\ \midrule
\multirow{2}{*}{Soil} & RMSE & 0.2522 $\pm$ 0.00 & 0.2609 $\pm$ 0.00 & - & - & - & - & - & - \\
 & SAD & 0.2097 $\pm$ 0.01 & 0.0959 $\pm$ 0.10 & - & - & - & - & - & 0.0328 \\
\multirow{2}{*}{Tree} & RMSE & 0.2614 $\pm$ 0.00 & 0.3431 $\pm$ 0.00 & - & - & - & - & - & - \\
 & SAD & 0.5347 $\pm$ 0.03 & 1.2788 $\pm$ 1.28 & - & - & - & - & - & 0.0519 \\
\multirow{2}{*}{Water} & RMSE & 0.2098 $\pm$ 0.00 & 0.3165 $\pm$ 0.00 & - & - & - & - & - & - \\
 & SAD & 0.8233 $\pm$ 0.04 & 0.4022 $\pm$ 0.40 & - & - & - & - & - & 0.1026 \\ \midrule
\multirow{2}{*}{Average} & RMSE & 0.2412 $\pm$ 0.00 & 0.3078 $\pm$ 0.00 & 0.0545 & 0.0239 & 0.0119 & 0.0006 & 0.0862 & 0.0171 \\
 & SAD & 0.5525 $\pm$ 0.03 & 0.5923 $\pm$ 0.59 & - & - & - & - & - & 0.0624 \\ \bottomrule
\end{tabular}%
}
\end{table*}

\begin{table*}[!h]
\centering
\caption{Abundance Estimation and Endmember Extraction Results on HYDICE Urban Dataset}
\label{table2}
\resizebox{\textwidth}{!}{%
\begin{tabular}{@{}ccccccccccc@{}}
\toprule
\multicolumn{2}{c}{} & SpACNN-LDVAE & MLP-LDVAE \cite{ldvae} & SSWNMF \cite{SSWNMF} & SGSNMF \cite{SGSNMF} & TV-RSNMF \cite{TV-RSNMF} & RSNMF \cite{TV-RSNMF} & GLNMF \cite{GLNMF} & $L_{1/2}$NMF \cite{L1/2NMF} & VCA+FCLS \cite{vca} \\ \midrule
\multirow{2}{*}{Asphalt road} & RMSE & 0.1566 $\pm$ 0.00 & 0.2889 $\pm$ 0.00 & - & - & - & - & - & - & - \\
 & SAD & 0.2786 $\pm$ 0.02 & 0.4262 $\pm$ 0.43 & 0.0782 $\pm$ 3.29 & 0.0841 $\pm$ 4.01 & 0.0770 $\pm$ 2.97 & 0.0869 $\pm$ 3.81 & 0.1008 $\pm$3.19 & 0.0889 $\pm$ 2.88 & 0.2246 $\pm$ 3.44 \\
\multirow{2}{*}{Grass} & RMSE & 0.1977 $\pm$ 0.00 & 0.1832 $\pm$ 0.00 & - & - & - & - & - & - & - \\
 & SAD & 0.1936 $\pm$ 0.01 & 0.3323 $\pm$ 0.33 & 0.1490 $\pm$ 3.58 & 0.1516 $\pm$ 3.25 & 0.1495 $\pm$ 3.54 & 0.1594 $\pm$ 3.62 & 0.1531 $\pm$ 3.06 & 0.1452 $\pm$ 3.57 & 0.1981 $\pm$ 3.39 \\
\multirow{2}{*}{Tree} & RMSE & 0.1632 $\pm$ 0.00 & 0.1737 $\pm$ 0.00 & - & - & - & - & - & - & - \\
 & SAD & 0.4411 $\pm$ 0.04 & 0.3177 $\pm$ 0.32 & 0.1173 $\pm$ 3.46 & 0.1199 $\pm$ 3.36 & 0.1269 $\pm$ 4.02 & 0.1457 $\pm$ 4.29 & 0.1424 $\pm$ 3.79 & 0.1509 $\pm$ 3.18 & 0.2137 $\pm$ 2.41 \\
\multirow{2}{*}{Roof} & RMSE & 0.1283 $\pm$ 0.00 & 0.125 $\pm$ 0.00 & - & - & - & - & - & - & - \\
 & SAD & 0.4502 $\pm$ 0.03 & 0.4393 $\pm$ 0.44 & 0.0713 $\pm$ 3.61 & 0.0731 $\pm$ 3.54 & 0.0746 $\pm$ 4.09 & 0.0849 $\pm$ 3.90 & 0.0986 $\pm$ 4.62 & 0.0863 $\pm$ 4.06 & 0.2673 $\pm$ 3.77 \\
\multirow{2}{*}{Metal} & RMSE & 0.0992 $\pm$ 0.00 & 0.2599 $\pm$ 0.00 & - & - & - & - & - & - & - \\
 & SAD & 0.3241 $\pm$ 0.02 & 0.7004 $\pm$ 0.70 & 0.1241 $\pm$ 2.76 & 0.1250 $\pm$ 3.81 & 0.1247 $\pm$ 3.53 & 0.1324 $\pm$ 4.15 & 0.1370 $\pm$ 4.28 & 0.1334 $\pm$ 3.90 & 0.1848 $\pm$ 3.68 \\
\multirow{2}{*}{Dirt} & RMSE & 0.1894 $\pm$ 0.00 & 0.1334 $\pm$ 0.00 & - & - & - & - & - & - & - \\
 & SAD & 0.2026 $\pm$ 0.01 & 0.2806 $\pm$ 0.28 & 0.0802 $\pm$ 3.17 & 0.0859 $\pm$ 3.91 & 0.0849 $\pm$ 3.92 & 0.0798 $\pm$ 3.77 & 0.1059 $\pm$ 3.96 & 0.1063 $\pm$ 3.54 & 0.1992 $\pm$ 3.43 \\ \midrule
\multirow{2}{*}{Average} & RMSE & 0.1558 $\pm$ 0.00 & 0.1840 $\pm$ 0.00 & 0.0048 $\pm$ 0.72 & 0.0061 $\pm$ 0.67 & 0.0055 $\pm$ 0.81 & 0.0053 $\pm$ 0.98 & 0.0069 $\pm$ 0.85 & 0.0044 $\pm$0.76 & 0.0119 $\pm$ 0.66 \\
 & SAD & 0.3151 $\pm$ 0.02 & 0.4161 $\pm$ 0.42 & 0.1034 $\pm$ 3.31 & 0.1060 $\pm$ 3.68 & 0.1063 $\pm$ 3.68 & 0.1148 $\pm$ 3.92 & 0.1230 $\pm$ 3.52 & 0.1185 $\pm$ 3.52 & 0.2142 $\pm$ 3.35 \\ \bottomrule
\end{tabular}%
}
\end{table*}

\begin{table*}[!h]
\centering
\caption{Endmember Extraction Results on Cuprite Dataset}
\label{table3}
\resizebox{\textwidth}{!}{%
\begin{tabular}{@{}cccccccccc@{}}
\toprule
 & SpACNN-LDVAE & MLP-LDVAE \cite{ldvae} & SSWNMF \cite{SSWNMF} & SGSNMF \cite{SGSNMF} & TV-RSNMF \cite{TV-RSNMF} & RSNMF \cite{TV-RSNMF} & GLNMF \cite{GLNMF} & $L_{1/2}$NMF \cite{L1/2NMF} & VCA+FCLS \cite{vca} \\ \midrule
alunite & 0.0683 $\pm$0.00 & 0.0097 $\pm$0.01 & 0.1497 $\pm$ 3.97 & 0.1238 $\pm$ 4.01 & 0.1204 $\pm$ 4.37 & 0.1189 $\pm$ 4.39 & 0.1353 $\pm$ 3.83 & 0.1496 $\pm$ 3.32 & 0.1574 $\pm$ 3.71 \\
Andradite & 0.0462 $\pm$0.00 & 0.0381 $\pm$0.04 & - & - & - & - & - & - & - \\
Buddingtonite & 0.0227 $\pm$0.00 & 0.0051 $\pm$0.01 & 0.0958 $\pm$ 4.69 & 0.1021 $\pm$ 3.47 & 0.0903 $\pm$ 5.08 & 0.1342 $\pm$ 4.72 & 0.1437 $\pm$ 3.62 & 0.1441 $\pm$ 4.16 & 0.1412 $\pm$ 3.74 \\
Dumortierite & 0.0500 $\pm$0.00 & 0.1922 $\pm$0.19 & - & - & - & - & - & - & - \\
Kaolinite\_1 & 0.0740 $\pm$0.00 & 0.0258 $\pm$0.03 & 0.0885 $\pm$ 2.94 & 0.0986 $\pm$ 3.18 & 0.1097 $\pm$ 3.47 & 0.0955 $\pm$ 3.07 & 0.0967 $\pm$ 4.01 & 0.0825 $\pm$ 4.66 & 0.0736 $\pm$ 4.42 \\
Kaolinite\_2 & 0.0249 $\pm$0.00 & 0.0699 $\pm$0.07 & 0.1206 $\pm$ 3.67 & 0.1375 $\pm$ 3.48 & 0.1213 $\pm$ 3.82 & 0.1396 $\pm$ 4.11 & 0.1356 $\pm$ 3.91 & 0.1402 $\pm$ 4.18 & 0.1420 $\pm$ 4.16 \\
Muscovite & 0.0320 $\pm$0.00 & 0.0064 $\pm$0.01 & 0.1024 $\pm$ 4.24 & 0.1061 $\pm$ 3.18 & 0.1131 $\pm$ 2.88 & 0.0997 $\pm$ 3.46 & 0.0961 $\pm$ 3.77 & 0.0889 $\pm$ 3.03 & 0.1007 $\pm$ 3.31 \\
Montmorillonite & 0.0214 $\pm$0.00 & 0.0496  $\pm$0.05 & 0.0651 $\pm$ 3.08 & 0.0705 $\pm$ 3.36 & 0.0783 $\pm$ 3.95 & 0.0744 $\pm$ 3.12 & 0.0838 $\pm$ 4.28 & 0.0876 $\pm$ 2.91 & 0.0974 $\pm$ 3.39 \\
Nontronite & 0.0639 $\pm$0.00 & 0.1048 $\pm$0.10 & 0.1138 $\pm$ 4.15 & 0.1046 $\pm$ 3.80 & 0.0911 $\pm$ 3.49 & 0.0832 $\pm$ 4.18 & 0.0953 $\pm$ 3.41 & 0.1038 $\pm$ 4.46 & 0.0772 $\pm$ 2.10 \\
Pyrope & 0.0342 $\pm$0.00 & 0.0156 $\pm$0.02 & 0.1106 $\pm$ 3.32 & 0.1208 $\pm$ 3.83 & 0.1253 $\pm$ 3.10 & 0.1469 $\pm$ 3.12 & 0.1318 $\pm$ 3.18 & 0.1123 $\pm$ 4.91 & 0.1437 $\pm$ 3.76 \\
Sphene & 0.1030 $\pm$0.00 & 0.0347  $\pm$0.03 & 0.1024 $\pm$ 3.79 & 0.1179 $\pm$ 4.02 & 0.1190 $\pm$ 2.97 & 0.1134 $\pm$ 2.54 & 0.1291 $\pm$ 4.21 & 0.1252 $\pm$ 5.18 & 0.1277 $\pm$ 4.08 \\
Chalcedony & 0.0281 $\pm$0.00 & 0.055  $\pm$0.01 & 0.1496 $\pm$ 4.12 & 0.1221 $\pm$ 4.02 & 0.1387 $\pm$ 4.01 & 0.1224 $\pm$ 4.19 & 0.1341 $\pm$ 2.98 & 0.1520 $\pm$ 3.43 & 0.1514 $\pm$ 3.83 \\ \midrule
Average & 0.0470 $\pm$0.00 & 0.0465 $\pm$0.05 & 0.1099 $\pm$ 3.80 & 0.1104 $\pm$ 3.63 & 0.1107 $\pm$ 3.71 & 0.1128 $\pm$ 3.69 & 0.1182 $\pm$ 3.72 & 0.1186 $\pm$ 4.02 & 0.1212 $\pm$ 3.65 \\ \bottomrule
\end{tabular}%
}
\end{table*}

\begin{table*}[!h]
\centering
\caption{Endmember Extraction Results on OnTech-Syn-HSI-21 Dataset}
\label{table4}
\resizebox{\textwidth}{!}{%
\begin{tabular}{@{}cccccccccc@{}}
\toprule
SNR & SpACNN-LDVAE & MLP-LDVAE \cite{ldvae} & SSWNMF \cite{SSWNMF} & SGSNMF \cite{SGSNMF} & TV-RSNMF \cite{TV-RSNMF} & RSNMF \cite{TV-RSNMF} & GLNMF \cite{GLNMF} & $L_{1/2}$NMF \cite{L1/2NMF} & VCA + FCLS \cite{vca} \\ \midrule
20 dB & 0.0584 $\pm$ 0.00 & 0.0224 $\pm$ 0.01 & 0.0636 $\pm$ 0.40 & 0.0782 $\pm$ 0.50 & 0.0679 $\pm$ 0.30 & 0.0731 $\pm$ 0.50 & 0.0724 $\pm$ 0.05 & 0.0744 $\pm$ 0.40 & 0.1358 $\pm$ 0.30 \\
30 dB & 0.0616 $\pm$ 0.00 & 0.0138 $\pm$ 0.01 & 0.0122 $\pm$ 0.01 & 0.0176 $\pm$ 0.03 & 0.0131 $\pm$ 0.03 & 0.0138 $\pm$ 0.05 & 0.0144 $\pm$ 0.04 & 0.0142 $\pm$ 0.04 & 0.0350 $\pm$ 0.06 \\
40 dB & 0.0613 $\pm$ 0.00 & 0.0081 $\pm$ 0.00 & 0.0029 $\pm$ 0.02 & 0.0033 $\pm$ 0.03 & 0.0036 $\pm$ 0.02 & 0.0041 $\pm$ 0.04 & 0.0044 $\pm$ 0.05 & 0.0037 $\pm$ 0.04 & 0.0125 $\pm$ 0.05 \\
50 dB & 0.0545 $\pm$ 0.00 & 0.0082 $\pm$ 0.00 & 0.0012 $\pm$ 0.02 & 0.0019 $\pm$ 0.02 & 0.0014 $\pm$ 0.03 & 0.0020 $\pm$ 0.04 & 0.0023 $\pm$ 0.04 & 0.0024 $\pm$ 0.03 & 0.0049 $\pm$ 0.06 \\
INF & 0.0594 $\pm$ 0.00 & 0.0069 $\pm$ 0.00 & - & - & - & - & - & - & - \\ \bottomrule
\end{tabular}%
}
\end{table*}

\begin{table*}[!h]
\centering
\caption{Abundance Estimation Results on OnTech-Syn-HSI-21 Dataset}
\label{table5}
\resizebox{\textwidth}{!}{%
\begin{tabular}{@{}cccccccccc@{}}
\toprule
SNR & SpACNN-LDVAE & MLP-LDVAE \cite{ldvae} & SSWNMF \cite{SSWNMF} & SGSNMF \cite{SGSNMF} & TV-RSNMF \cite{TV-RSNMF} & RSNMF \cite{TV-RSNMF} & GLNMF \cite{GLNMF} & $L_{1/2}$NMF \cite{L1/2NMF} & VCA + FCLS \cite{vca} \\ \midrule
20 dB & 0.0948 $\pm$ 0.00 & 0.0052 $\pm$ 0.00 & 0.1339 $\pm$ 0.20 & 0.1322 $\pm$ 0.40 & 0.1342 $\pm$ 0.30 & 0.1426 $\pm$ 0.40 & 0.1434 $\pm$ 0.60 & 0.1430 $\pm$ 0.50 & 0.1704 $\pm$ 0.03 \\
30 dB & 0.3356 $\pm$ 0.00 & 0.0302 $\pm$ 0.00 & 0.0386 $\pm$ 0.20 & 0.0391 $\pm$ 0.30 & 0.0420 $\pm$ 0.20 & 0.0426 $\pm$ 0.30 & 0.0429 $\pm$ 0.03 & 0.0432 $\pm$ 0.20 & 0.0548 $\pm$ 0.20 \\
40 dB & 0.3343 $\pm$ 0.00 & 0.0303 $\pm$ 0.00 & 0.0122 $\pm$ 0.03 & 0.0148 $\pm$ 0.05 & 0.0142 $\pm$ 0.04 & 0.0147 $\pm$ 0.05 & 0.0150 $\pm$ 0.04 & 0.0153 $\pm$ 0.03 & 0.0164 $\pm$ 0.10 \\
50 dB & 0.3335 $\pm$ 0.00 & 0.0303 $\pm$ 0.00 & 0.0041 $\pm$ 0.02 & 0.0059 $\pm$ 0.05 & 0.0050 $\pm$ 0.03 & 0.0055 $\pm$ 0.03 & 0.0064 $\pm$ 0.04 & 0.0061 $\pm$ 0.04 & 0.0087 $\pm$ 0.08 \\
INF & 0.0948 $\pm$ 0.00 & 0.0052 $\pm$ 0.00 & - & - & - & - & - & - & - \\ \bottomrule
\end{tabular}%
}
\end{table*}

\clearpage
\bibliographystyle{IEEEbib}
\bibliography{refs}
\end{document}